\documentclass[nohyperref]{article}

\usepackage{microtype}
\usepackage{graphicx}
\usepackage{subfigure}
\usepackage{booktabs} 
\usepackage{algorithm}
\usepackage{multirow}
\usepackage{hyperref}

\usepackage[accepted]{icml2022}

\usepackage{amsmath}
\usepackage{amssymb}
\usepackage{mathtools}
\usepackage{amsthm}

\usepackage{listings}
\usepackage{color}

\definecolor{dkgreen}{rgb}{0,0.6,0}
\definecolor{gray}{rgb}{0.5,0.5,0.5}
\definecolor{mauve}{rgb}{0.58,0,0.82}

\usepackage{pifont}
\newcommand{\xmark}{\ding{55}}%

\usepackage[capitalize,noabbrev]{cleveref}

\theoremstyle{plain}

\theoremstyle{definition}

\theoremstyle{remark}

\usepackage[textsize=tiny]{todonotes}

\icmltitlerunning{Tied-Augment: Controlling Representation Similarity}

\begin{document}

\twocolumn[
\icmltitle{Tied-Augment: Controlling Representation Similarity Improves \\ Data Augmentation}




\begin{icmlauthorlist}
\icmlauthor{Emirhan Kurtulus}{s,c}
\icmlauthor{Zichao Li}{g}
\icmlauthor{Yann Dauphin}{g}
\icmlauthor{Ekin D. Cubuk}{g}
\end{icmlauthorlist}

\icmlaffiliation{s}{Stanford University}
\icmlaffiliation{c}{Cagaloglu Anadolu Lisesi}
\icmlaffiliation{g}{Google Research, Brain Team}

\icmlcorrespondingauthor{Emirhan Kurtulus}{emirhank@stanford.edu}
\icmlcorrespondingauthor{Ekin D. Cubuk}{cubuk@google.com}

\icmlkeywords{Machine Learning, ICML}

\vskip 0.3in
]



\printAffiliationsAndNotice{}  

\begin{abstract}
Data augmentation methods have played an important role in the recent advance of deep learning models, and have become an indispensable component of state-of-the-art models in semi-supervised, self-supervised, and supervised training for vision. Despite incurring no additional latency at test time, data augmentation often requires more epochs of training to be effective. For example, even the simple flips-and-crops augmentation requires training for more than 5 epochs to improve performance, whereas RandAugment requires more than 90 epochs. We propose a general framework called Tied-Augment, which improves the efficacy of data augmentation in a wide range of applications by adding a simple term to the loss that can control the similarity of representations under distortions. Tied-Augment can improve state-of-the-art methods from data augmentation (e.g. RandAugment, mixup), optimization (e.g. SAM), and semi-supervised learning (e.g. FixMatch). For example, Tied-RandAugment can outperform RandAugment by 2.0\% on ImageNet. Notably, using Tied-Augment, data augmentation can be made to improve generalization even when training for a few epochs and when fine-tuning. We open source our code at \url{https://github.com/ekurtulus/tied-augment/tree/main}
\end{abstract}

\section{Introduction}
\label{sec:intro}

Data augmentation is an integral part of training deep neural networks to improve their performance by modulating the diversity and affinity of data \cite{aug_tradeoffs}. Although data augmentation offers significant benefits \cite{simard2003best,krizhevsky2017imagenet,shorten2019survey,szegedy2015going}, as the complexity of the augmentation increases, so does the minimum number of epochs required for its effectiveness \cite{autoaugment}. As neural networks and datasets get larger, machine learning models get trained for fewer epochs (for example, ~\citet{dosovitskiy2020image} pretrained for 7 epochs), typically due to computational limitations. In such cases, conventional data augmentation methods lose their effectiveness. Additionally, data augmentation is not as effective when finetuning pretrained models as it is when training from scratch. 

In this work, we present a general framework that mitigates these problems and is applicable to a range of problems from supervised training to semi-supervised learning by amplifying the effectiveness of data augmentation through feature similarity modulation. Our framework, Tied-Augment, makes forward passes on two augmented views of the data with tied (shared) weights. In addition to the classification loss, we add a similarity term to enforce invariance between the features of the augmented views. We find that our framework can be used to improve the effectiveness of both simple flips-and-crops (Crop-Flip) and aggressive augmentations even for few-epoch training. As the effect of data augmentation is amplified, the sample efficiency of the data increases. Therefore, our framework works well even with small amounts of data, as shown by our experiments on CIFAR-4K (4k samples from CIFAR-10), Oxford-Flowers102, and Oxford-IIT Pets.

Despite the simplicity of our framework, Tied-Augment empowers augmentation methods such as Crop-Flip and RandAugment~\cite{cubuk2020randaugment} to improve generalization even when trained for a few epochs, which we demonstrate for a diverse set of datasets. For longer training, Tied-Augment leads to significant improvements over already-strong baselines such as RandAugment and mixup~\cite{zhang2017mixup}. For example, Tied-RandAugment achieves a 2\% improvement over RandAugment when training ResNet-50 for 360 epochs on ImageNet, without any architectural modifications or additional regularization.
\clearpage
Our contributions can be summarized as follows:
\begin{itemize} 
    \item We show that adding a simple loss term to modulate feature similarity can significantly improve the effectiveness of data augmentation, which we show for a diverse set of data augmentations such as Crop-Flip, RandAugment, and mixup.
    \item Unlike conventional methods of data augmentation, with our framework, data augmentation can improve performance even when training for only a single epoch for finetuning pretrained networks or training from scratch on a wide range of datasets with different architectures.
    \item We compare Tied-Augment to multi-stage self-supervised learning methods (first pretraining, then finetuning on ImageNet). Our proposed framework is designed to be as straightforward as traditional data augmentation techniques, while avoiding the need for additional components such as a memory bank, large batch sizes, contrastive data instances, extended training periods, or large model sizes.
    Despite this simplicity, Tied-Augment can outperform more complex self-supervised learning methods on ImageNet validation accuracy.
\end{itemize}
\section{Background / Related Work}
\label{sec:related}
\subsection{Data Augmentation}
Data augmentation has been a critical component of recent advances in deep vision models \cite{he2022masked,bai2022improving,liu2021swin}. We can divide data augmentation works into two categories: individual operations and optimal combinations of individual operations. In general, data augmentation operations are performed to expand the distribution of the input space and improve performance. 

Random cropping and horizontal flips are widely used operations in image processing problems. This set of operations is usually extended by color operations \cite{szegedy2016rethinking,szegedy2017inception}. mixup \cite{zhang2017mixup} is a method that uses a convex sum of images and their labels. This operation provides better generalization and robustness even in the presence of corrupted labels. Other operations include Cutout \cite{devries2017improved}, a method that randomly masks out square regions within the image; Patch Gaussian \cite{lopes2019improving}, an operation that combines Cutout with the addition of Gaussian noise by randomly adding noise to selected square regions; \cite{liu2016ssd}, a cropping strategy for object detection that generates smaller training samples by taking crops from an upscaled version of the original image; Copy-Paste \cite{ghiasi2021simple}, an augmentation method that inserts random objects onto the selected training sample.

\subsection{Self-supervised Learning}
Self-supervised learning is a form of representation learning that usually makes use of pretext tasks to learn general representations \cite{ericsson2022self}. Generally, self-supervised learning methods follow a two-step paradigm. They first pretrain the network on a large dataset, then use it for fine-tuning on downstream tasks. 

Clustering is the paradigm of mapping non-linear augmented views projections into a unit sphere of K classes \cite{bautista2016cliquecnn}. This paradigm is notably widespread in image understanding problems \cite{caron2018deep,asano2019self,caron2019unsupervised,gidaris2020learning}. SwAV \cite{caron2020unsupervised} is particularly noteworthy in this set of works. They cluster the data by enforcing consistency between the assigned clusters of the augmented views. Additionally, they propose \textit{multi-crop} strategy, a random cropping strategy that not only two standard resolution crops but also N low resolution crops to take the features of varying resolutions into account.

Contrastive instance discrimination learns representations by pushing the features of positive instances, meaning augmented views of the same image or images with same classes, closer and pushing features of negative instances away \cite{hadsell2006dimensionality}. Currently, this is one of the most widely used paradigms.

MoCo \cite{he2020momentum} maintains a dictionary of encodings and views the problem as query matching. SimSiam \cite{chen2021exploring} proposes to encode two augmented views of the same image, one with an MLP (multi-layer perceptron) head, and increase feature similarity. BYOL \cite{grill2020bootstrap} follows the same method, but uses a network and another network following it by exponential moving average. SimCLR \cite{chen2020simple} uses a network with an MLP head to encode two augmented views and maximizes similarity through contrastive loss \cite{hadsell2006dimensionality}. NNCLR \cite{dwibedi2021little} improves on this approach by using clustering to maximize the number of correct negative instances. SupCon \cite{khosla2020supervised} adapts this paradigm to supervised learning by following SimCLR and using contrastive loss, but selecting the correct positive and negative labels by using labels. SupCon showed that augmentation methods such as RandAugment with a supervised-contrastive loss can outperform the same data augmentation methods with a cross-entropy loss.

\section{Tied-Augment}
\label{sec:method}
\begin{figure}[!t]
\begin{minipage}[t]{\linewidth}
\hfill
\end{minipage}
\begin{minipage}[t]{\linewidth}
\begin{lstlisting}
# model: a neural network that returns features and logits
# tw: tied-weight
# augment1: a stochastic data augmentation module
# augment2: a stochastic data augmentation module
# note that augment1 can be the same as augment2

# ce = cross entropy loss
# mse = mean squared error loss
   
for x,y in loader:
   # generate two augmented views of the same image
   x1 = augment1(x)
   x2 = augment2(x)

   # extract features and logits 
   f1, l1 = model(x1)
   f2, l2 = model(x2)
      
   # calculate loss
   ce_loss = (ce(l1, y) + ce(l2, y)) / 2
   feature_loss = mse(f1, f2) 
   loss = ce_loss + tw * feature_loss

\end{lstlisting}
\caption{Python code for Tied-Augment based on NumPy.}
\label{fig:algo}
\end{minipage}
\end{figure}

\begin{figure}
    \centering
    \includegraphics[width=\linewidth]{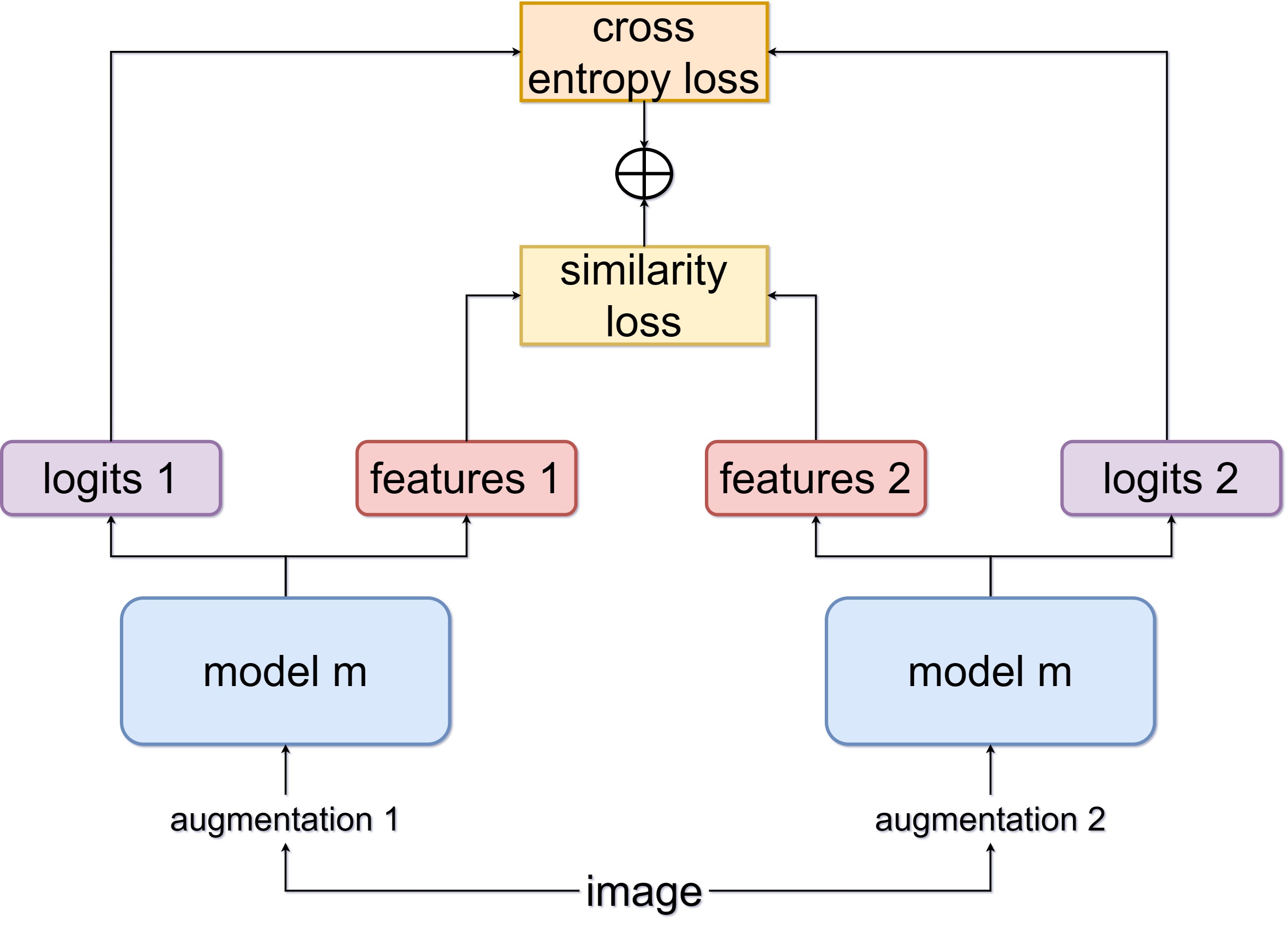}
    \caption{Tied-Augment framework.}
    \label{fig:framework}
\end{figure}

Tied-Augment framework combines supervised and representation learning in a simple way. We propose to enforce pairwise feature similarity between given augmented views of the same image while using a supervised loss signal. As shown in Figure \ref{fig:framework}, our framework consists of three components:
\begin{itemize} \item \textbf{Two stochastic data augmentation modules} (can be identical) produce two augmented views of the same image. These transformations can be chosen arbitrarily as long as they improve the performance of the baseline supervised model. However, in this work, we use the same augmentation for both branches for simplicity. Given two augmentations, we name the case after the more complex augmentation. For example, if RandAugment is used with Crop-Flip on the other branch, we name the case Tied-RandAugment. In Section \ref{sec:experiments} we provide a thorough analysis of the effects of the chosen data augmentation modules. 
\item \textbf{A neural network} generates features (pre-logits) and logits based on given an image. There are no architectural constraints as our framework is based on the pre-logit feature vector, which is used in all classification networks. 
\item \textbf{Pairwise feature similarity and supervised loss functions} enforce pairwise feature similarity/dissimilarity and supervised loss signal, respectively. In this work, we use L2 loss as the pairwise feature similarity function (we ablate this decision in \ref{sec:ablations}) and, for simplicity, cross entropy loss as the supervised loss. The contribution of the feature-similarity term to the loss is controllable by the hyperparameter Tied-weight.
\end{itemize}

The training of Tied-Augment works as follows. In each training iteration, we sample a batch of images of size $N$, and generate two augmented views, resulting in $2N$ images. For each image pair, we compute the feature similarity with L2 loss and for each image we calculate cross entropy loss. For given input $x$, logits $f(x)$, labels $y$, features of the first augmented views $v_{1}=v_1(x)$, features of the second augmented views $v_{2}=v_2(x)$, supervised loss $\ell$, and the feature similarity loss weight $w$ the loss function of Tied-Augment is:
\begin{align}
    \mathcal{L}_\text{Tied-Aug}= \sum_i\ell(f(v_i(x)), y) + w \|v_1(x) - v_2(x)\|^2
\end{align}
In Algorithm \ref{fig:algo}, we provide an overview of our framework. In general, the views correspond directly to the feature representations $v_i=h_i=h(\text{aug}_i(x))$ where $h$ is the function that produces the feature representation and $\text{aug}_i(\cdot)$ is the $i$-th augmentation function. However, we will also examine cases that require more elaborate views such as Tied-mixup.

\subsection{Tied-FixMatch}
In this section, we apply the Tied-Augment framework to FixMatch \cite{sohn2020fixmatch} as a case study to demonstrate the easy adaptability of our framework. We refer to this version as Tied-FixMatch. FixMatch is a semi-supervised learning algorithm that combines consistency regularization and pseudo-labeling. For the labeled portion of the dataset, FixMatch uses standard cross-entropy loss. For the unlabeled images, FixMatch generates two augmented views of the same image using a weak (Crop-Flip) and a strong (RandAugment) transformation. Then, the model's predictions for the weakly augmented image are used as pseudo-labels for the strongly-augmented image. However, predictions whose confidence is below a threshold are masked out and not used to calculate the unsupervised loss. 
FixMatch uses a standard cross-entropy loss denoted $\ell_s$ on the labeled images.
%
%

Considering that FixMatch already has a two-branch strategy for learning from unlabeled images, we can introduce Tied-Augment without any additional computational cost. In Tied-FixMatch, we change the objective as not only maximizing consistency and minimizing pseudo-labeling loss, but also minimizing the pairwise feature distance between augmented views of the same unlabeled images. In doing so, we also mask the instances with a confidence threshold and do not apply the pairwise feature similarity loss. Therefore, given features of the weakly-augmented unlabeled images $h_{1}$, features of the strongly-augmented unlabeled images $h_{2}$, and a similarity loss weight $w$, the loss minimized by Tied-FixMatch is simply:
\begin{align}
    \ell_s + \lambda_u \ell_u + w \|h_{1} - h_{2}\|^{2}
\end{align}

\subsection{Tied-mixup}

Here, we consider the application of Tied-Augment to mixup \citep{zhang2017mixup}. mixup is a popular data augmentation technique that produces augmented examples by convex combination of pairs of training points
\begin{align}
    \hat{x}  &= \lambda x_1 + (1-\lambda) x_2 \nonumber\\
    \hat{y}  &= \lambda y_1 + (1-\lambda) y_2 \nonumber
\end{align}
where $\lambda \sim \text{Beta}(\alpha, \alpha)$ is a mixing coefficient sampled from a Beta distribution with parameter $\alpha$.

Unlike the previously considered augmentations, different mixup augmented views have different labels in general. Applying Tied-Augment to mixup requires defining a better correspondence between the two augmented views.
We propose the following
\begin{align}
    \Omega_M(h)=w\|\lambda h(x_1) + (1-\lambda)h(x_2) - h(\hat{x})\|^2.
\end{align}
 In order to produce features that are in the same space as the first view of mixed examples $v_1=h(\hat{x})$, this approach mixes the features of the clean examples to produce the second view $v_2=\lambda h(x_1) + (1-\lambda)h(x_2)$. In effect, this is encouraging the features of the model to be linear in-between training points.

\subsection{Tied-SAM (Sharpness Aware Minimization)}
Sharpness-Aware Minimization (SAM) \cite{foret2020sharpness} is a widely-used training strategy that consists of two steps. At the first step, SAM applies an adversarial perturbation to first place the weights at the highest point in the loss landscape. Then, in the second step, this results in a move to a wider minimum. Tied-SAM augments this algorithm by boosting the adversarial move through pushing features of the augmented views apart (negating the Tied-weight) in the first step. In doing so, enable SAM to find a better adversarial loss landscape location. For the second step, we apply standard Tied-Augment to move to an even wider minimum.

\subsection{Understanding Tied-Augment}
\label{sec:analysis}




We can gain some insight into TiedAugment by considering its application to Gaussian input noise augmentation. The additive regularization for Tied-GaussianNoise  is given by
\begin{equation} 
\Omega_G(f) = w E_{\epsilon}[\|h({\bf x}) - h({\bf x + \epsilon})\|^2]\nonumber
\end{equation}
where $h$ produces the features of the network, ${\bf x}\in \mathbb{R}^n$, and $\epsilon \sim \mathcal{N}(0, \sigma)^n$. Consider the approximation of this term using the first-order Taylor expansion of $h$ at ${\bf x}$,
$\Omega_G(h) \approx w \sigma^2 \|\nabla h({\bf x})\|^2_F.$
This additive regularization is part of the well known class of Tikhonov regularizers \citep{tikhonov1977solutions,bishop1995training} that include weight decay. It encourages the feature mapping function to become more invariant to small corruptions in the input, which can be beneficial for generalization. For a more detailed analysis of Tied-Augment, please refer to Appendix \ref{appendix:theory}.

\section{Experiments}
\label{sec:experiments}
To show the effectiveness of Tied-Augment, we experiment with training from-scratch on CIFAR-10, CIFAR-100, and ImageNet. We extend these tests with finetuning and few-epoch / low-data regimes to simulate more realistic scenarios, where the amount of domain-specific data or available compute is limited. Lastly, we show that Tied-Augment significantly improves the performance of state-of-the-art methods (e.g. mixup and SAM) and can be used for semi-supervised learning (e.g. FixMatch). For all models that use RandAugment, we show its configuration as ``RandAugment(N=number of layers, M=magnitude, P=probability)''. If probability is not given, it is set to the default of 1.0.


\subsection{CIFAR-10 and CIFAR-100, CIFAR-4K}
CIFAR-10 and CIFAR-100 are widely studied datasets, and CIFAR-4K is a benchmark intended to simulate the low-data regime. All baselines and Tied-Augment models include random pad-and-crops and flips (CF). RandAugment baselines and Tied-RandAugment models also include Cutout \cite{devries2017improved}. For RandAugment experiments, we copy the reported optimal number of layers (N) and magnitude (M) for both augmentation branches, decoupling the hyperparameter search space from augmentation selection. We did not find an additional improvement from tuning the RandAugment on the second branch (e.g. RandAugment(N=2, M=14) for one branch, and RandAugment(N=2, M=19) for the second). We also experimented with Stacked-RandAugment \cite{tian2020makes} and SimAugment \cite{chen2020simple} on the second branch but saw no performance improvement over standard RandAugment. On both CIFAR-10 and CIFAR-100, we use the same data augmentation pairs for Wide-ResNet-28-10 and Wide-ResNet-28-2. All models are trained using the hyperparameters from RandAugment \cite{cubuk2020randaugment}. 

Additionally, we measure the efficacy of Tied-Augment on CIFAR-4K. We randomly sample 400 images from each class for training and leave the test set as is. We use the same hyperparameters as \citet{cubuk2020randaugment} including training for 500 epochs. We use the same optimal setting of RandAugment(N=2, M=14) on both branches. As shown in Table \ref{tab:CIFAR-10_results}, Tied-Augment improves both Crop-Flip and RandAugment by a significant amount, on all CIFAR datasets considered. We report all the hyperparameters in Appendix \ref{appendix:cifar}.

\begin{table}[]\small\centering
\begin{tabular}{lcc|cc}
                                       & CF & Tied-CF & RA & Tied-RA \\ \hline
\multicolumn{1}{l|}{\textbf{CIFAR-10}} & \multicolumn{1}{l}{}         & \multicolumn{1}{l|}{}        & \multicolumn{1}{l}{}         & \multicolumn{1}{l}{}        \\
\multicolumn{1}{l|}{WRN-28-2}            & 94.9 & \textbf{95.5} & 95.8 & \textbf{96.9} \\
\multicolumn{1}{l|}{WRN-28-10}           & 96.1 & \textbf{96.5} & 97.3 & \textbf{98.1} \\ \hline
\multicolumn{1}{l|}{\textbf{CIFAR-100}}  &      &               &      &               \\
\multicolumn{1}{l|}{WRN-28-2}            & 75.4 & \textbf{76.9} & 79.3 & \textbf{80.4} \\
\multicolumn{1}{l|}{WRN-28-10}           & 81.2 & \textbf{81.6} & 83.3 & \textbf{85.0} \\ \hline
\multicolumn{1}{l|}{\textbf{CIFAR-4K}} &      &               &      &               \\
\multicolumn{1}{l|}{WRN28-2}             & 82.0 & \textbf{82.5} & 85.3 & \textbf{87.8} \\
\multicolumn{1}{l|}{WRN28-10}            & 83.5 & \textbf{84.5} & 86.8 & \textbf{90.2} \\ \hline
\end{tabular}
\caption{\textbf{Test accuracy (\%) on CIFAR-10, CIFAR-100, CIFAR-4K, and ImageNet datasets.} We compare Tied-Augment to Crop-Flip (CF) and RandAugment (RA) baselines. Reported results are the average of 5 independent runs. Standard deviation of the results for each experiment is smaller than or equal to 0.1\%.}
\label{tab:CIFAR-10_results}
\end{table}

\begin{table}[]\scriptsize\centering
\begin{tabular}{lc|c|cc|cc}
 &
  \#epochs &
  \multicolumn{1}{l|}{Identity} &
  \begin{tabular}[c]{@{}c@{}}Baseline\\ (CF)\end{tabular} &
  \begin{tabular}[c]{@{}c@{}}Tied-\\ CF\end{tabular} &
  \begin{tabular}[c]{@{}c@{}}Baseline\\ (RA)\end{tabular} &
  \begin{tabular}[c]{@{}c@{}}Tied-\\ RA\end{tabular} \\ \hline
\multirow{4}{*}{CIFAR-10}  & 1  & 72.6 & 70.3 & \textbf{73.3} & 72.8 & \textbf{73.4} \\
                          & 2  & 76.1 & 75.1 & \textbf{82.8} & 81.4 & \textbf{82.4} \\
                          & 5  & 89.2 & 88.5 & \textbf{89.5} & 88.6 & \textbf{89.5} \\
                          & 10 & 91.8 & 91.8 & \textbf{92.2} & 91.0 & \textbf{92.5} \\ \hline
\multirow{4}{*}{CIFAR-100} & 1  & 26.9 & 24.6 & \textbf{28.1} & 22.8 & \textbf{29.4} \\
                          & 2  & 41.4 & 39.4 & \textbf{42.6} & 41.7 & \textbf{43.9} \\
                          & 5  & 62.4 & 60.6 & \textbf{62.8} & 61.7 & \textbf{62.2} \\
                          & 10 & 70.8 & 70.4 & \textbf{71.2} & 71.2 & \textbf{71.7} \\ \hline
\end{tabular}
\caption{Test accuracy (\%) for few-epoch training on CIFAR datasets. Reported results are the average of 10 independent runs. For 1, 2, 5, 10 epochs, standard deviations are below 0.5, 0.3, 0.2, and 0.1 respectively.}
\label{tab:few_epoch}
\end{table}

\begin{table}[h!]\small\centering
    \begin{tabular}{c|c|cccc}
    \#epochs &
      \multicolumn{1}{l}{Identity} &
      \begin{tabular}[c]{@{}c@{}}Baseline\\ CF\end{tabular} &
      \multicolumn{1}{c|}{\begin{tabular}[c]{@{}c@{}}Tied-\\ CF\end{tabular}} &
      \begin{tabular}[c]{@{}c@{}}Baseline \\ (RA)\end{tabular} &
      \begin{tabular}[c]{@{}c@{}}Tied-\\ RA\end{tabular} \\ \hline
    \textbf{Cars}     & \multicolumn{1}{l}{} & \multicolumn{1}{l}{} & \multicolumn{1}{l}{}              & \multicolumn{1}{l}{} & \multicolumn{1}{l}{} \\
    2                 & 69.0                 & 59.9                 & \multicolumn{1}{c|}{\textbf{69.5}} & 58.7                 & \textbf{69.4}        \\
    5                 & 80.9                 & 81.6                 & \multicolumn{1}{c|}{\textbf{84.7}} & 81.4                 & \textbf{84.6}        \\
    10                & 82.0                 & 86.7                 & \multicolumn{1}{c|}{\textbf{88.3}} & 87.1                 & \textbf{89.2}        \\
    25                & 82.0                 & 88.9                 & \multicolumn{1}{c|}{\textbf{89.4}} & 90.4                 & \textbf{91.5}        \\
    50                & 82.3                 & 89.6                 & \multicolumn{1}{c|}{\textbf{90.0}} & 91.5                 & \textbf{92.2}        \\ \hline
    \textbf{Flowers}  & \multicolumn{1}{l}{} & \multicolumn{1}{l}{} & \multicolumn{1}{l}{}               & \multicolumn{1}{l}{} & \multicolumn{1}{l}{} \\
    2                 & 56.6                 & 47.1                 & \multicolumn{1}{c|}{\textbf{56.8}} & 47.2                 & \textbf{56.5}        \\
    5                 & 88.3                 & 86.4                 & \multicolumn{1}{c|}{\textbf{88.7}} & 84.7                 & \textbf{88.7}        \\
    10                & 90.7                 & 91.6                 & \multicolumn{1}{c|}{\textbf{93.3}} & 92.1                 & \textbf{93.5}        \\
    25                & 91.8                 & 93.9                 & \multicolumn{1}{c|}{\textbf{94.1}} & 93.5                 & \textbf{94.3}        \\
    50                & 92.2                 & 93.6                 & \multicolumn{1}{c|}{\textbf{94.5}} & 94.1                 & \textbf{95.1}        \\ \hline
    \textbf{Pets}     & \multicolumn{1}{l}{} & \multicolumn{1}{l}{} & \multicolumn{1}{l}{}               & \multicolumn{1}{l}{} & \multicolumn{1}{l}{} \\
    2                 & 91.4                 & 91.4                 & \multicolumn{1}{c|}{\textbf{92.1}} & 91.4                 & \textbf{92.0}        \\
    5                 & 92.4                 & 92.8                 & \multicolumn{1}{c|}{\textbf{93.1}} & 92.1                 & \textbf{93.0}        \\
    10                & 92.5                 & 93.1                 & \multicolumn{1}{c|}{\textbf{93.3}} & 92.9                 & \textbf{93.2}        \\
    25                & 92.9                 & 93.4                 & \multicolumn{1}{c|}{\textbf{93.7}} & 93.4                 & \textbf{93.6}        \\
    50                & 92.8                 & 93.5                 & \multicolumn{1}{c|}{\textbf{93.8}} & 93.5                 & \textbf{93.7}        \\ \hline
    \textbf{Aircraft} & \multicolumn{1}{l}{} & \multicolumn{1}{l}{} & \multicolumn{1}{l}{}               & \multicolumn{1}{l}{} & \multicolumn{1}{l}{} \\
    2                 & 44.2                 & 34.1                 & \multicolumn{1}{c|}{\textbf{41.8}} & 31.6                 & \textbf{40.8}        \\
    5                 & 58.2                 & 51.1                 & \multicolumn{1}{c|}{\textbf{58.3}} & 50.6                 & \textbf{58.1}        \\
    10                & 59.3                 & 60.6                 & \multicolumn{1}{c|}{\textbf{61.9}} & 60.7                 & \textbf{61.5}        \\
    25                & 61.2                 & 68.8                 & \multicolumn{1}{c|}{\textbf{69.9}} & 72.3                 & \textbf{74.6}        \\
    50                & 62.3                 & 71.6                 & \multicolumn{1}{c|}{\textbf{72.3}} & 74.2                 & \textbf{76.1}        \\ \hline
    \textbf{CIFAR-10} & \multicolumn{1}{l}{} & \multicolumn{1}{l}{} & \multicolumn{1}{l}{}               & \multicolumn{1}{l}{} & \multicolumn{1}{l}{} \\
    2                 & 95.7                 & 95.2                 & \multicolumn{1}{c|}{\textbf{95.9}} & 95.1                 & \textbf{95.9}        \\
    5                 & 96.4                 & 96.3                 & \multicolumn{1}{c|}{\textbf{96.8}} & 96.3                 & \textbf{96.8}        \\
    10                & 96.5                 & 96.8                 & \multicolumn{1}{c|}{\textbf{97.1}} & 96.8                 & \textbf{97.2}        \\
    25                & 96.6                 & 97.2                 & \multicolumn{1}{c|}{\textbf{97.4}} & 97.3                 & \textbf{97.6}        \\
    50                & 96.6                 & 97.2                 & \multicolumn{1}{c|}{\textbf{97.4}} & 97.6                 & \textbf{97.8}        \\ \hline
    \end{tabular}
    \caption{\textbf{Finetuning experiments on Stanford-cars, Oxford-Flowers102 (Flowers), Oxford-IIIT Pets (Pets), FGVC Aircraft (Aircraft), CIFAR-10 datasets.} Reported results for the 2, 5 and 10 epoch experiments are the average of 10 independent runs, while the rest is the average of 5 independent runs. Baseline results are the maximum of standard training and double augmentation branch with no similarity loss. pretrained model is a standard ResNet-50, Tied-Augment is only used for finetuning. Best CF (CF vs. Tied-CF) and RA (RA vs. Tied-RA) results are bolded. The standard deviations of the accuracies are smaller than or equal to 0.5\%, 0.4\%, 0.2\%, 0.1\%, and 0.1\% for 2, 5, 10, 25, and 50 epochs respectively.}
    \label{tab:finetuning}
    \end{table}

\subsection{Few-epoch training}
Previous work has shown that data augmentation is only able to improve generalization when the model is trained for more than a certain number of epochs. Usually, more complex data augmentation protocols require more epochs. For example, \citet{autoaugment} reported that more than 90 epochs was required to be able to search and apply AutoAugment policies. Similarly, ~\citet{lopes2019improving} reported that none of the tested augmentation transformations was helpful when trained for only 1 epoch, even for simple augmentations such as flips or crops. To test how much of this problem can be mitigated by Tied-Augment, we evaluate our method on CIFAR-10 and CIFAR-100 for \{1, 2, 5, 10\} epochs. For runs with epoch=\{1, 2, 5\}, the learning rate and weight-decay were tuned to maximize the validation accuracy of the identity baseline (since in this regime identity baseline outperforms the Crop-Flip baseline). The learning rate and weight-decay hyperparameters for the 10 epoch models were tuned to maximize the validation set performance of the Crop-Flip baseline. 

To ensure fairness by eliminating the possibility of doubled epochs introduced by the two forward passes of our framework, in all reported tasks, the baselines performances are the maximum of standard training (no similarity loss and single augmentation branch) and double augmentation branch (with variable augmentation methods) with no similarity loss. Unlike our 200 epochs CIFAR experiments, we do not use the same augmentation for both branches but allow both the baseline and the Tied-Augment model to combine any one of the following augmentation methods: RandAugment(N=1, M=2), RandAugment(N=2, M=14), Crop-Flip, and identity. If one of the branches use RandAugment, for instance RandAugment for one branch and identity for the other, then it is only compared to RandAugment runs.

In Table \ref{tab:few_epoch}, we show that Tied-Augment can outperform identity transformation in the epoch regimes as small as 2. Unconventionally, Tied-Augment is capable of pushing RandAugment to the level of Crop-Flip and identity, and even outperforming them in the \{2, 5, 10\} epochs regimes. For all the epoch regimes, Tied-Augment outperforms its baseline significantly, up to 6.7\%.

In addition to training networks from scratch for a limited number of epochs, finetuning for a few epochs is also an important problem given the ever-growing trend to go deeper for neural networks. Therefore, we test our framework on finetuning tasks where data augmentation is considerably less effective than from-scratch training. For this purpose, we train a ImageNet-pretrained ResNet-50 \cite{he2016deep} model on Stanford-Cars \cite{krause20133d}, Oxford Flowers \cite{nilsback2008automated}, Oxford Pets \cite{parkhi2012cats}, FGVC Aircraft \cite{maji13fine-grained}, and CIFAR-10 \cite{krizhevsky2009learning}. Table \ref{tab:finetuning} compares the performance of our framework to the baselines models. It is evident that, like our from-scratch experiments, Tied-Augment is able to outperform identity with not only a weak augmentation like Crop-Flip but with RandAugment. On all the finetuning datasets we experimented with, Tied-Augment consistently and significantly improves the baseline, up to 10.7\%.


\subsection{Image classification on ImageNet}\label{sec:imagenet}
We train ResNet-50 and ResNet-200 architectures~\cite{he2016deep} on the ImageNet dataset~\cite{deng2009imagenet} with RandAugment. Previous work had shown that aggressive data augmentation strategies such as AutoAugment or RandAugment do not improve generalization if trained only for 90 epochs. To see if Tied-Augment can fix this issue, we train with Tied-RandAugment on ResNet-50 for 90 epochs. To see the full benefit of aggressive data augmentation, we also train Tied-RandAugment models for 360 epochs. Finally, to see the impact our approach on simple augmentations, we train the standard ResNet-50 with the standard Crop-Flip baseline and our Tied-CropFlip. Finally, to test the impact of Tied-Augment on a larger model, we train ResNet-200 for 180 epochs. We train ResNet-200 for fewer epochs to compensate for its larger compute requirement. We do not observe an improvement on the baseline ResNet-200 models when training for longer. All ResNet models use the standard training hyperparameters for ResNet, listed in Appendix Section \ref{appendix:imagenet}.

In Table~\ref{tab:imagenet}, we find that Tied-RandAugment is able to improve top-1 accuracy by almost 2\% when trained for 90 epochs, and significantly reduces the number of epochs required for RandAugment to be effective, whereas regular RandAugment requires more than 90 epochs to improve generalization. When trained for 360 epochs, Tied-RandAugment still improves on RandAugment by 2\%, totalling a 3.3\% improvement over simple Crop-Flip. We also observe that Tied-CropFlip outperforms regular Crop-Flip in every setting.

To evaluate Tied-Augment on a different data augmentation method, we trained Resnet-50 networks with the same setup with mixup. We cross-validate the mixup coefficient $\alpha$ within the values $\{0.2, 0.3, 0.4\}$, and the similarity loss weight within $\{1, 50, 100\}$. Our mixup baseline achieves an top-1 accuracy of 77.9\%. When we apply our simple Tied-Augment framework to mixup, Tied-mixup achieves 78.8\%, an almost 1\% improvement over an already strong baseline. 

Since the Tied-Augment loss has a supervised and an unsupervised term, we compare Tied-Augment to relevant self-supervised methods that utilize all the training labels of ImageNet in addition to self-supervised training on ImageNet samples. We find that even though Tied-RandAugment is trained for fewer epochs without the need for multiple stages of training, Tied-RandAugment outperforms other methods for both ResNet-50 and ResNet-200 (Table~\ref{tab:self_supervised}). 

\begin{table}[]\small
    \begin{tabular}{lc|cc|cc}
    \multicolumn{1}{c}{\textbf{ImageNet}} & \#epochs & CF & Tied-CF & RA & Tied-RA \\ \hline
    ResNet-50  & 90  & 76.3 & \textbf{77.0} & 76.3 & \textbf{78.2} \\
    ResNet-50  & 360 & 76.3 & \textbf{76.9} & 77.6 & \textbf{79.6} \\ \hline
    ResNet-200 & 180 & 78.5 & \textbf{79.7} & 80.0 & \textbf{81.8} \\ \hline
    \end{tabular}
\caption{\textbf{ImageNet results.} CF and RA refer to Crop-Flip and RandAugment, respectively. ResNet-200 baselines do not improve when trained for more than 180 epochs. Standard deviations for the reported results are smaller than or equal to 0.2\%.}
\label{tab:imagenet}
\end{table}
\begin{table}[]
\begin{tabular}{l|cc|c}
             & Epochs & Multi-stage & Top-1 \\ \hline
\textbf{ResNet-50} &&& \\
SimCLR       & 1000   & \checkmark   & 76.0  \\ 
SimCLR v2    & 800    &\checkmark    & 76.3  \\ 
BYOL         & 1000   & \checkmark   & 77.7  \\ 
SupCon       & 350    & \checkmark   & 78.7  \\ 
Tied-RandAugment & 360    & \xmark       & \textbf{79.6}  \\ \hline
\textbf{ResNet-200} &&& \\
SupCon       & 700    & \checkmark   & 81.4  \\ 
Tied-RandAugment & 360    & \xmark       & \textbf{81.8}  \\ 
\hline
\end{tabular}
\caption{\textbf{Comparison of our method to self-supervised models.} Multi-stage denotes the need for separate pretraining and finetuning stages. Note that Tied-Augment methods do not require a pretraining stage. Performances of the self-supervised models are their 100\% ImageNet finetuned results. Results reported are the average of 5 independent runs. The standard deviations are smaller than or equal to 0.2\% for all reported results.}
\label{tab:self_supervised}
\end{table}

\subsection{Transferability of Tied-Augment Features}
We finetune a Tied-ResNet50 to downstream datasets to measure the transferability of its features and compare it to BYOL \cite{grill2020bootstrap}, SimCLR-v2 \cite{chen2020big}, and SwAV \cite{caron2020unsupervised}. We follow the SSL-Transfer \cite{ericsson2021well} framework for our finetuning experiments. Namely, we finetune for 5000 step using a batch size of 64, SGD with Nesterov momentum \cite{sutskever2013importance}, doing a grid search over learning rate and weight decay. We choose the learning rate from 4 logarithmically spaced values between 0.0001 and 0.1. Weight decay is chosen from 4 logarithmically spaced values between $10^{-6}$ and $10^{-3}$ as well as 0.

In Table \ref{tab:ssl_finetune}, we compare the performance of Tied-Augment to self-supervised models and the supervised baseline. Our model outperforms SwAV \cite{caron2020unsupervised} by 0.8\% and the supervised baseline by 1.6\%. This shows that the features learned by Tied-Augment are more transferable than its self-supervised and supervised counterparts. It is worth noting that a Tied-RandAugment model finetuned using Tied-CropFlip significantly improves an already strong performacee (0.9\%).

\begin{table*}[]\small\centering
\begin{tabular}{c|cccccccccc|c}
\hline
\textbf{}  & Aircraft & Cal-101 & Cars  & CIFAR-10 & CIFAR-100 & DTD   & Flowers & Food           & Pets  & SUN397         & Avg   \\ \hline
SimCLR v2  & 78.7    & 82.9      & 79.8 & 96.2   & 79.1    & 70.2 & 94.3   & 82.2          & 83.2 & 61.1          & 80.8 \\
BYOL       & 79.5    & 89.4      & 84.6 & 97.0   & 84.0    & 73.6 & 94.5   & 85.5          & 89.6 & 64.0          & 84.2 \\
SwAV       & 83.1    & 89.9      & 86.8 & 96.8   & 84.4    & 75.2 & 95.5   & \textbf{87.2} & 89.1 & \textbf{66.2} & 85.4 \\
Supervised & 83.5    & 91.0      & 82.6 & 96.4   & 82.9    & 73.30 & 95.5   & 84.6          & 92.4 & 63.6          & 84.6 \\
\begin{tabular}[c]{@{}c@{}}Tied-RA \end{tabular} &
  84.7 &
  92.6 &
  89.9 &
  96.9 &
  83.9 &
  75.8 &
  96.7 &
  84.3 &
  93.5 &
  63.9 &
  86.2 \\ \hline
\begin{tabular}[c]{@{}c@{}}Tied-RA +\\Tied-CF finetune\end{tabular} &
  \textbf{88.1} &
  \textbf{93.3} &
  \textbf{90.2} &
  \textbf{97.2} &
  \textbf{85.2} &
  \textbf{76.2} &
  \textbf{97.3} &
  86.4 &
  \textbf{93.9} &
  64.5 &
  \textbf{87.1} \\ \hline
\end{tabular}
\caption{Finetuning experiments on downstream datasets comparing self-supervised learning to Tied-Augment pretrained model. All reported models are ResNet50. Supervised baseline is pretrained using only RandAugment. SimCLR-v2, BYOL, SwAV, and supervised baseline are from \cite{ericsson2021well}. Tied-RA stands for Tied-RandAugment. Tied-RA + Tied-CF finetune is the case where a Tied-RA pretrained ResNet50 is finetuned using Tied-CropFlip. All models are finetuned using crop-flip data augmentation.}
\label{tab:ssl_finetune}
\end{table*}

\subsection{Tied-FixMatch}
To back up our claim that we offer a framework that can be used for a wide-range of problems, we apply Tied-Augment to a semi-supervised learning algorithm: FixMatch \cite{sohn2020fixmatch}. We compare the performance of our framework to the baseline exactly following the hyperparameters of the original work, without changing the augmentation pair of the unsupervised branch or adding the similarity term to the supervised branch. We use Wide-ResNet-28-2 and Wide-ResNet-28-8 configurations for CIFAR-10 and CIFAR-100 respectively. For the unsupervised branch, we use Crop-Flip for the weak branch and RandAugment(N=2, M=10, probability=0.5) for the strong branch, while supervised branch uses Crop-Flip. For our CIFAR-10 and CIFAR-100 experiments, we use 4000 and 10000 labeled examples preserving the class balance respectively. In Table \ref{tab:fixmatch}, it is shown that Tied-FixMatch not only outperforms the baseline FixMatch but also outperforms its supervised counterpart which uses all of the 50000 labeled images. All hyperparameters are listed in Appendix \ref{appendix:fixmatch}.

\begin{table}[]\centering
\begin{tabular}{l|c|c|c}
 &
  \begin{tabular}[c]{@{}c@{}}FixMatch\\ baseline\end{tabular} &
  \begin{tabular}[c]{@{}c@{}}Supervised\\ baseline\end{tabular} &
  \begin{tabular}[c]{@{}c@{}}Tied-\\ FixMatch\end{tabular} \\ 
  \hline
  \#labels & 4k & 50k & 4k \\
  \hline
CIFAR-10 &
  95.7 $\pm$ 0.05 &
  95.8 $\pm$ 0.02 &
  \textbf{96.1} $\pm$ 0.04 \\
CIFAR-100 &
  77.4 $\pm$ 0.12 &
  77.6 $\pm$ 0.04 &
  \textbf{77.9} $\pm$ 0.08 \\ \hline
\end{tabular} 
\caption{\textbf{Application of Tied-Augment framework to FixMatch.} Similarity function is applied to the features between the unsupervised branches. The reported FixMatch baseline results are from \cite{sohn2020fixmatch}, supervised baseline results are from \cite{cubuk2020randaugment} and include RandAugment, and our results are the average of 5 runs.}
\label{tab:fixmatch}
\end{table}

\subsection{Composability of Tied-Augment}
It is crucial for a framework to be composable with other methods while retaining their performance improvements. To show that Tied-Augment has this property, we experiment with Sharpness-Aware Minimization \cite{foret2020sharpness}. For SAM experiments, we train a Wide-ResNet-28-10 following the hyperparameters of the original work for 200 epochs which are listed in Appendix \ref{appendix:sam}. We replicate their results with RandAugment(N=2, M=14). In Table \ref{tab:sam}, we show that Tied-SAM outperforms the baseline SAM.

\begin{table}[]\centering
\begin{tabular}{l|c|c|c}
         & \begin{tabular}[c]{@{}c@{}}Supervised\\ baseline\end{tabular} & \begin{tabular}[c]{@{}c@{}}SAM\\ baseline\end{tabular} & Tied-SAM      \\ \hline
CIFAR-10  & 97.3  {\scriptsize $\pm$ 0.03}                                                     & 97.9  {\scriptsize $\pm$ 0.1}                                                 & \textbf{98.3} {\scriptsize $\pm$ 0.1} \\ \hline
CIFAR-100 & 83.3 {\scriptsize $\pm$ 0.05}                                                         & 86.2 {\scriptsize $\pm$ 0.1}                                                   & \textbf{86.5} {\scriptsize $\pm$ 0.1} \\ \hline
\end{tabular}
\caption{\textbf{Sharpness-Aware minimization (SAM) experiments.} Baselines are replicated. Supervised baseline and SAM baseline both include RandAugment. The reported results are the average of 5 independent runs.}
\label{tab:sam}
\end{table}
\section{Ablations and Analysis}
\label{sec:ablations}
In this section, we analyze the components of Tied-Augment framework and show their effectiveness. Additionally, we ablate our design choices.
\begin{table*}[h!]
    \centering 
    \textbf{Different components of Tied-Augment}
    \begin{tabular}{lrrrr}
        \hline
         & ImageNet & CIFAR-10 & Stanford-Cars  \\
         &  & & (2 epochs) \\
        \hline
        (1) Baseline (Flips and Crops) &76.3 & 96.1 & 59.9 \\
        (2) RandAugment &77.6 & 97.3 & 58.7 \\
        (3) Two views with same RandAugment policy &\multirow{1}{*}{78.0} & \multirow{1}{*}{97.6}& \multirow{1}{*}{52.4}\\
        (4) Two views with different RandAugment policies &78.5 & 97.7 & 54.2 \\ 
        (5) Tied-Augment &\textbf{79.6} & \textbf{98.1}  &  \textbf{69.4}\\
    \end{tabular}
    \caption{\textbf{Ablation study for the improvements coming from Tied-Augment on ImageNet, CIFAR-10, and CIFAR-100.} Relative to a baseline model, addition of two augmented views of the same image improves performance (3). Creating two augmented views by two distinct augmentation methods (generally one more aggressive RandAugment, and one less aggressive RandAugment) further boosts performance (4). Finally, adding a feature similarity objective yields a significant performance increase (5).}
    \label{tab:tied_components}
\end{table*}
\subsection{Deconstructing Tied-Augment}
In Table~\ref{tab:tied_components}, we deconstruct Tied-Augment framework and show the improvement from each component. For each task considered, we create the highest-performing Tied-Augment method by first starting with the simplest baseline (standard crop-flip). Then, we apply RandAugment. Even though RandAugment provides noteworthy performance benefits (e.g. 1.3\% on ImageNet), it is not effective and even harmful for finetuning and few-epoch training. Since Tied-Augment requires two differently augmented views of a sample, some of its improvement comes from ``augmenting the batch''~\citep{hoffer2020augment,fort2021drawing} (row (3)). We find additional benefits from diversifying the augmentation policies used for the different views (row (4)). Finally, the largest improvement comes from ``tying'' the representations coming from the two branches, which gives us Tied-RandAugment (row (5)), which adds an additional 1.1\%, 0.4\%, and 15.2\% accuracy on ImageNet, CIFAR-10, and Stanford-Cars (epochs), respectively, in addition to our improved diversely augmented batch approach. 

We find that for few-epoch from-scratch and fning experiments, generally 2 or 5 epochs, supervised signal from only one branch shows a better performance. In other cases, however, we found that cross entropy loss on both batches $b_1 \text{ and } b_2$ improves the results more. 

We, then, discuss the computational cost Tied-Augment below. Tied-Augment requires a single forward pass and a single backward pass. If there is no I/O bottleneck and a high-end accelerator (e.g Nvidia A100), the runtime of a forward pass on $b_1$ is roughly equal to a forward pass on $b_1 \text{ and } b_2$. However, from a number of computational operations perspective, the required computation is double the forward pass of standard training. The cost of the backward pass on $b_1 \text{ and } b_2$ size is approximately the same as a backward pass on $b_1$ on modern accelerators. Therefore, Tied-Augment only increases the computational cost by the additional forward pass; however, it is still computationally cheaper than double-step methods like SAM because it does not require two separate backward passes. For example, instead of a 100\% increase in computational cost (as would be the case for SAM), we empirically observe an increase of roughly 30\% increase on an Nvidia A100 on CIFAR-10. 

\subsection{Similarity Function}
One of the critical components of Tied-Augment is the similarity term. In Table \ref{tab:similarity_ablation}, we report the results of L1, L2 and cosine similarity functions. Here, it is worth noting that in the reported results, the weight of the cosine function is negative unlike L1 and L2 in the sense that for maximizing feature similarity L1 and L2 need to be minimized while cosine angle between the representations needs to be maximized. It is a known phenomenon that data augmentation can improve the model output invariance to distortions~\cite{aug_tradeoffs}. Therefore, it is intuitive to also encourage representation invariance. Interestingly, we find that the opposite can also be true. Enforcing feature dissimilarity can also improve performance on highly overparametrized CIFAR datasets considered; however, this is not the case for ImageNet for L2 similarity function. For simplicity (halving the search space for Tied-weight) and maximum performance improvement on all considered datasets, we choose only to consider increasing invariance. It is worth noting that negative Tied-weights for L1 and L2 (minimizing feature similarity) on CIFAR datasets also outperforms the baseline (Tied-weight=0). For cosine similarity, positive Tied-weight can outperform baseline for all datasets considered. We provide an analysis of the stability of tied-weight in Appendix \ref{appendix:stability}.   

\subsection{General Design Choices}
In Tied-Augment framework, there are many design choices that are of interest. For example, given that we double the batch size, there are two ways of doing the forward pass: separate forward passes on the batches or a single forward pass on both of the batches concatenated. 
These two approaches are not functionally equivalent for networks with BatchNorm (BN) \cite{ioffe2015batch} due to the running statistics.
We find that the performance difference between these cases is generally equal to or less than 0.1\%. We consistently report the results of double separate forward passes. 

Another design choice to consider is the use of BN layers. For our experiments where we use two different RandAugment configurations (one weak, one stronger), we evaluated Split BatchNorms~\cite{xie2020adversarial,merchant2020does} but did not find significant performance improvements. Thus we only report experiments that use standard BN layers. 

Being invariant to different crops is a desirable property when targeting occlusion-invariance~\cite{purushwalkam2020demystifying}. We also try using the same crop for both branches in our Tied-RandAugment experiments. This means taking a random (resized for ImageNet) crop from the image once and feeding the same crop into RandAugment on both augmentation branches. Surprisingly, this has little to no effect on performance. Therefore, for simplicity, we use different crops on both augmentation branches for CIFAR and finetuning experiments, same crop for ImageNet experiments. 

\begin{table}[]
\begin{tabular}{c|c|c|c}
 & \begin{tabular}[c]{@{}c@{}}similarity\\ function\end{tabular} & \begin{tabular}[c]{@{}c@{}}Tied-\\ Crop-Flip\end{tabular} & Tied-RA \\ \hline
\multirow{3}{*}{CIFAR-10}  & L1     & 96.3          & 97.8          \\
                           & Cosine & \textbf{96.5} & 98.0          \\
                           & L2     & \textbf{96.5} & \textbf{98.1} \\ \hline
\multirow{3}{*}{CIFAR-100} & L1     & 81.3          & 84.8          \\
                           & Cosine & 81.5          & \textbf{85.0} \\
                           & L2     & \textbf{81.6} & \textbf{85.0} \\ \hline
\multirow{3}{*}{ImageNet}  & L1     &  \textbf{76.9} & 78.7         \\
                           & Cosine &  76.7 & 78.8              \\
                           & L2     &  \textbf{76.9}& \textbf{79.2}  \\ \hline
\end{tabular}
\caption{\textbf{Ablation on the similarity function.} Tied-weights of all considered similarity functions have the signs so that they increase the feature similarity. Reported results are the average of 5 distinct runs. Imagenet Tied-RA models use (N=2, M=9) on both branches.} 
\vspace{-0.3cm}
\label{tab:similarity_ablation}
\end{table}

\begin{figure}
    \centering
    \includegraphics[width=\linewidth]{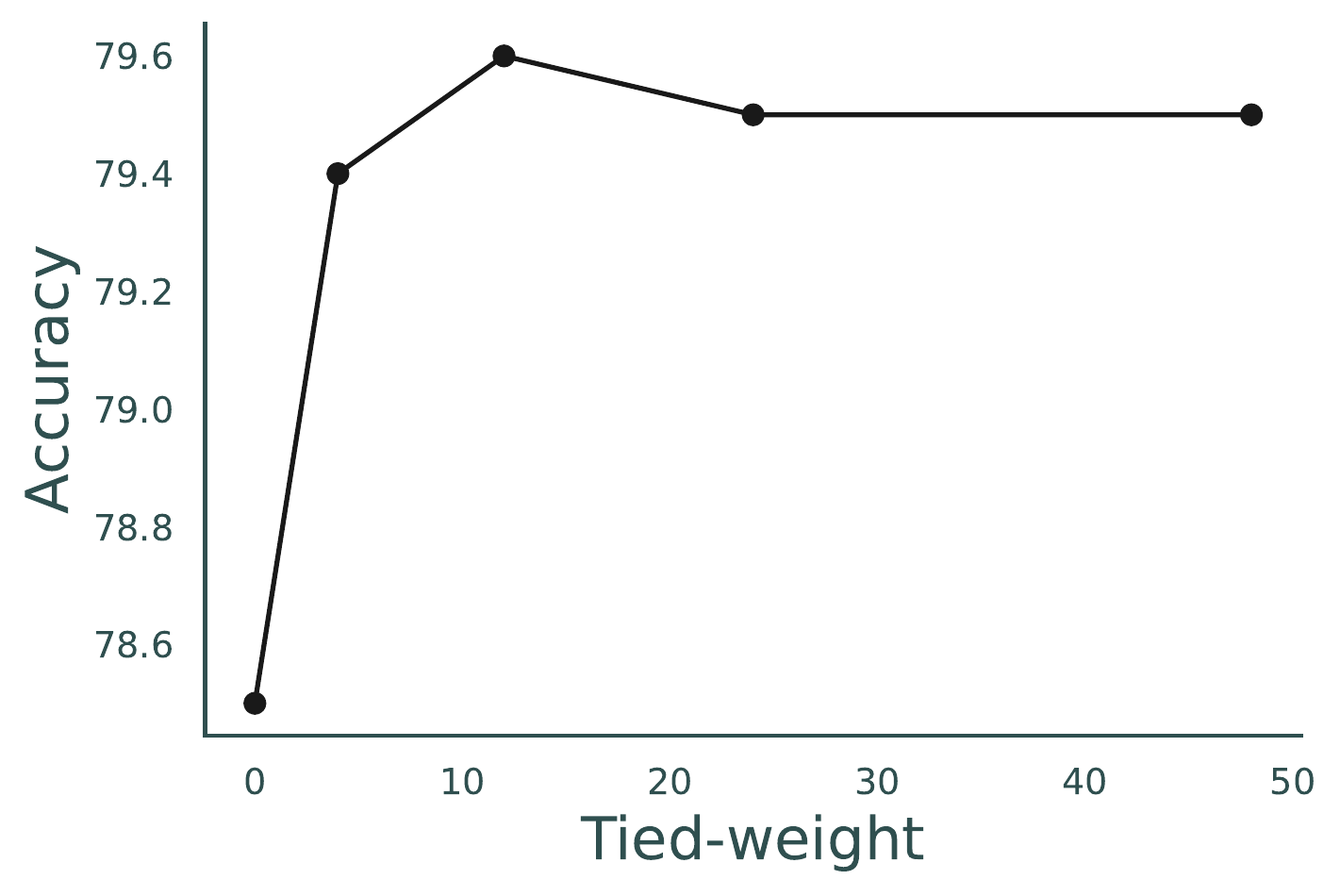}
    \caption{Tied-RandAugment accuracy with ResNet-50 on ImageNet as a function of tied-weight.}
    \label{fig:stability}
\end{figure}

\subsection{Stability of tied-weight}\label{appendix:stability}
In Figure \ref{fig:stability}, we present the stability of the introduced tied-weight hyperparameter. It is shown that even for a large range of values, tied-weight is able to improve the performance of the model on ImageNet dataset, indicating that Tied-Augment offers significant performance improvements without the need for extensive hyperparameter search.
\section{Conclusion}
\label{sec:conclusion}

As dataset and model sizes increase, machine learning models are trained for fewer and fewer epochs. Traditionally this has made data augmentation less useful. We introduce Tied-Augment, a simple method for combining self-supervised learning and regular supervised learning to strengthen state-of-the-art methods such as mixup, SAM, FixMatch, and RandAugment by up to 2\% on Imagenet. Tied-Augment can be implemented with only a few lines of additional code. 

Tied-Augment can improve the effectiveness of standard data augmentation approaches such as Crop-Flip even when training for a few epochs. When training for longer, Tied-Augment achieves significant improvements over state-of-the-art augmentation methods. 

Tied-Augment shows the promise of combining self-supervised approaches with regular supervised learning. An exciting future direction would be to evaluate Tied-Augment for large language model training which tends to be for a few epochs.
\clearpage
\section{Acknowledgments}
We thank Omer Faruk Ursavas for his contributions to this project. The numerical calculations reported in this paper were partially performed at TUBITAK ULAKBIM, High Performance and Grid Computing Center (TRUBA resources). We acknowledge the support of CuRe program from Google AI, Jonathan Caton, and the Google Cloud team. We acknowledge Johannes Gasteiger for his feedback on the manuscript, Jascha Sohl-dickstein for helpful discussions. 

\bibliography{references.bib}
\bibliographystyle{icml2022}
\clearpage
\section{Appendix}
\label{section:hparams}

\subsection{ImageNet}\label{appendix:imagenet}
All ImageNet models use a learning rate of 0.4 with a batch size of 1024, weight-decay rate of 1e-4. The Tied-RandAugment model that was trained for 90 epochs used Crop-Flip on first branch, and RandAugment(N=2, M=9) on the other branch, with a Tied-weight of 4. The Tied-RandAugment ResNet-50 model that was trained for 360 epochs used RandAugment(N=2, M=13) for the first branch and RandAugment(N=2, M=9, P=0.5) for the second branch, with a Tied-weight of 12.0. The Tied-RandAugment ResNet-200 model used RandAugment(N=2, M=13) for both branches with a Tied-weight of 12.0. 

All Tied-Augment ImageNet models trained for 90 epochs used a Tied-weight of 4, and models trained for longer used a Tied-weight of 12. The optimal Tied-weight for Tied-mixup on Imagenet was 50.

\subsection{CIFAR-10, CIFAR-100 and CIFAR-4K}\label{appendix:cifar}
For CIFAR-4K, we use a learning rate of 0.1 with a batch size of 128 and weight-decay rate of 5e-4. All the baselines and Tied-Augment models were trained for 500 epochs. The reported models for Wide-ResNet-28-10 and Wide-ResNet-28-2 configurations use RandAugment(N=2, M=14) for both branches with a Tied-weight of 10 and 24 respectively. 

CIFAR-10 and CIFAR-100 models use a learning rate of 0.1 with a batch size of 128 and weight-decay rate of 5e-4 and were trained for 200 epochs. For both Wide-ResNet-28-2 and Wide-ResNet-10 configurations, we use RandAugment(N=2, M=14) for the first branch and RandAugment(N=2, M=19) for the second branch. Tied-weights for the reported results for Wide-ResNet-28-2 and Wide-ResNet-28-10 are 16 and 20 respectively.

\subsection{Tied-FixMatch}\label{appendix:fixmatch}
For our Tied-FixMatch experiments on CIFAR-10 and CIFAR-100, we set the FixMatch parameters as follows: $\tau$=0.95, $\lambda_u$=1, $\mu$=7, batch size=64, learning rate=0.03 using SGD with Nesterov momentum. We set weight decay to 0.0005 and 0.001 for CIFAR-10 and CIFAR-100 respectively.

\subsection{Tied-SAM}\label{appendix:sam}
On CIFAR-10 and CIFAR-100, we use learning rate of 0.1, batch size of 256, weight decay of 0.0005. SAM hyperparameter $\rho$ to 0.05 and 0.1 for CIFAR-10 and CIFAR-100 respectively.

\subsection{Comparison between Self-Supervised Learning and Tied-Augment}\label{appendix:selfsupervised}
Given that Tied-Augment combines self-supervised learning with supervised learning, it is important to understand the intuition behind this framework. One intuitive observation is that purely self-supervised methods can sometimes suffer from representation-collapse which is converging to the trivial solution of outputting zeros for all inputs, which would make sure representations are the same for differently augmented samples. This could also happen if we only trained with our similarity loss.

This intuition seems to be relevant to why self-supervised training can be unstable and papers have focused on removing this instability. For example, SimCLR has added an additional layer for contrastive learning that gets discarded during finetuning. Other self-supervised methods such as MoCo had to use momentum and large batch sizes to stabilize training. In our case, since we use supervision from the beginning, the representation-collapse or other instabilities do not occur. It is of course possible to increase the tied-weight sufficiently to cause collapse, but in practice a simple hyperparameter search over values of \{1, 5, 10, 50\} is sufficient and tied-weight up to 10 is stable for most experimental setups. In addition to increasing performance in a stable way without the need for search over many hyperparameters or other tricks required self-supervised training, Tied-Augment provides a significant improvement in performance even with the very same hyperparameters with its supervised baseline. Therefore, in the presence of labels, Tied-Augment is more favorable than supervised or self-supervised learning.

One important advantage of self-supervised learning is its promise of not requiring labels in which case Tied-Augment cannot be used. We would like to draw attention to the fact that, as evinced by Tied-FixMatch, Tied-Augment is beneficial even if there are only few labels available. Additionally, existence of methods like CLIP and large language models and datasets such as LAION-5B \cite{schuhmann2022laion} and JFT-300 \cite{sun2017revisiting} show us that (1) supervision is possible without real class labels (2) curating large datasets that are noisly and weakly labeled is possible and such datasets are extremely effective. This shows us that the combination of supervision and self-supervision will be a crucial paradigm in the future which we propose in this paper.

\subsection{More Detailed Tied-Augment Analysis}\label{appendix:theory}
 The following is the analysis sketch for various approaches
 We are tying together two different augmentation distributions $P(\tilde{\bf x}_1, \tilde{\bf y}_1|{\bf x},{\bf y}), P(\tilde{\bf x}_2, \tilde{\bf y}_2|{\bf x},{\bf y})$ as follows
 \begin{align}
     R_\text{TiedAug}(f, h) = \frac{1}{N}\sum_{({\bf x},{\bf y})} E [\ell(f(h(\tilde{\bf x}_1)), \tilde{\bf y}_1) + \\ \|h(\tilde{\bf x}_1) - m(h(\tilde{\bf x}_2))\|^2] 
 \end{align}
 where ${\bf x}\in \mathbb{R}^n$ is the input, ${\bf y}\in \mathbb{R}^m$ are the labels, $f$ is the final classifier on the features provdided by $h$. $m$ is a function that ensures that the hidden features from both transformations correspond to the same class
 \begin{align}
     P({\bf y}|h(\tilde{\bf x}_1)) = P({\bf y}|m(h(\tilde{\bf x}_2)))
 \end{align}
 In the case of augmentations that do not change the class such as additive gaussian noise, the identity function will suffice $m(x)=x$.
 \subsubsection{Tied-GaussianNoise with L2 distance}

We can gain some insight into TiedAugment by considering its application to Gaussian input noise augmentation. The additive regularization for Tied-GaussianNoise  is given by
\begin{equation} 
\Omega_G(f) = w E_{\epsilon}[\|h({\bf x}) - h({\bf x + \epsilon})\|^2]\nonumber
\end{equation}
where $h$ produces the features of the network, ${\bf x}\in \mathbb{R}^n$, and $\epsilon \sim \mathcal{N}(0, \sigma)^n$. Consider the approximation of this term using the first-order Taylor expansion of $h$ at ${\bf x}$,
$\Omega_G(h) \approx w \sigma^2 \|\nabla h({\bf x})\|^2_F.$
 \begin{align} 
 \Omega_G(h) &\approx w E_{{\bf x}, \epsilon}[\|\nabla_{\bf x}f({\bf x})^T\epsilon\|^2] \nonumber\\
 &= w E_{\bf x}[\nabla_{\bf x}^Tf({\bf x})E_\epsilon[\epsilon\epsilon^T]\nabla_{\bf x}h({\bf x})] \nonumber\\
 &= w \sigma^2 E_{\bf x}[\|\nabla_{\bf x}f({\bf x})\|^2_F]
 \end{align}
This additive regularization is part of the well known class of Tikhonov regularizers \citep{tikhonov1977solutions,bishop1995training} that include weight decay. It encourages the feature mapping function to become more invariant to small corruptions in the input, which can be beneficial for generalization.

 \subsubsection{(worse approximation) TiedGaussianNoise with cosine similarity }
 The regularization for TiedGaussianNoise with cosine similarity is given by
 \begin{equation} 
 \Omega_{TGNCS}(h) = \lambda E\left[\frac{h({\bf x})^Th({\bf x} + \epsilon)}{\|h({\bf x})\| \|h({\bf x} + \epsilon)\|}\right]
 \end{equation}
 Consider the second-order Taylor expansion of h around $\bf x$
 \begin{equation} 
 \approx \lambda E\left[\frac{h({\bf x})^T(h({\bf x}) + \nabla_{\bf x}h({\bf x})\epsilon + \epsilon^T\nabla^2_{\bf x}h({\bf x})\epsilon)}{\|h({\bf x})\| \|h({\bf x} + \epsilon)\|}\right]
 \end{equation}
 Now, consider the first-order Taylor expansion of the norm
 \begin{equation} 
 E[\|h({\bf x} + \epsilon)\|] \approx \|h({\bf x})\| + \nabla_{\bf x}(\|h({\bf x})\|)E[\epsilon]
 \end{equation}
 Given that the noise is zero-mean the second term disappears
 \begin{equation} 
 E[\|h({\bf x} + \epsilon)\|] \approx E[\|h({\bf x})\|]
 \end{equation}
 Putting this together we have
 \begin{equation} 
 \approx \lambda E\left[\frac{h({\bf x})^T(h({\bf x}) + \nabla_{\bf x}h({\bf x})\epsilon + \epsilon^T\nabla^2_{\bf x}h({\bf x})\epsilon)}{\|h({\bf x})\|^2}\right]
 \end{equation}
 which simplifies to
 \begin{equation} 
 = \lambda + 0 + \lambda E\left[\frac{h({\bf x})^T\epsilon^T\nabla^2_{\bf x}h({\bf x})\epsilon}{\|h({\bf x})\|^2}\right]
 \end{equation}
 dropping the constant term and simplifying further we have
 \begin{equation} 
 \lambda \sum_i \frac{h_i({\bf x})}{\|h({\bf x})\|}\frac{\text{Tr}(\nabla^2_{\bf x}h_i({\bf x}))}{\|h({\bf x})\|}
 \end{equation}

\end{document}